\pdfoutput=1
\documentclass[10pt, logo, twocolumn, copyright, nonumbering]{nvidiatechreport}

\usepackage{hyperref}
\usepackage{url}
\usepackage{colortbl}
\usepackage{nicefrac}       %
\usepackage{microtype}      %
\usepackage[dvipsnames]{xcolor}         %
\usepackage{multirow}
\usepackage{multicol}
\usepackage{graphicx}
\usepackage{xspace}
\usepackage{amsmath}
\usepackage{adjustbox}
\usepackage{tcolorbox}
\usepackage{amssymb}
\usepackage{enumitem}
\usepackage{wrapfig}
\usepackage{xcolor}
\usepackage{subcaption}    %
\usepackage{float}
\usepackage{stfloats}
\usepackage{amsmath}
\usepackage{listings}
\usepackage{mdframed}
\usepackage{tcolorbox}
\tcbuselibrary{most}
\usepackage{arydshln}
\usepackage{booktabs} 
\usepackage{color,soul}
\usepackage{makecell}
\usepackage[numbers]{natbib}
\usepackage{caption}
\usepackage{makecell}
\usepackage{tabularx}
\usepackage{float}
\usepackage{placeins}

\tcbuselibrary{listings,breakable}

\definecolor{pastelblue}{RGB}{173,216,230}
\definecolor{pastelyellow}{RGB}{255,253,208}
\definecolor{pastelpink}{RGB}{255,209,220}
\definecolor{pastelgreen}{RGB}{176,226,172}
\definecolor{pastellavender}{RGB}{230,230,250}

\definecolor{NvidiaGreen}{RGB}{118, 185, 0}
\sethlcolor{red!15}

\usepackage{xcolor}         
\usepackage{booktabs}
\usepackage[table]{xcolor} 

\newcommand{\qwenOneSevenB}{\texttt{Qwen3-1.7B}\xspace}

\newtcblisting{llmprompt}[1][]{
  enhanced,
  breakable,
  colback=promptbg,
  colframe=promptbd,
  coltext=prompttl,
  boxrule=0.5pt,
  arc=3mm,
  left=10pt,right=10pt,top=10pt,bottom=10pt,
  shadow={0.5mm}{-0.5mm}{0mm}{black!10},
  listing only,
  listing options={
    basicstyle=\small\ttfamily,
    columns=fullflexible,
    showstringspaces=false,
    keepspaces=true,
    upquote=true,
    breaklines=true,
    breakatwhitespace=false
  },
  title={#1},
  colbacktitle=black!6,
  coltitle=prompttl,
  fonttitle=\bfseries\large,
  attach boxed title to top left={yshift=-2pt, xshift=8pt},
  boxed title style={
    sharp corners,
    colframe=promptbd,
    boxrule=0.5pt,
    interior style={fill=black!6},
    left=6pt,right=8pt,top=4pt,bottom=4pt
  }
}

\title{\centering{Learning Generative Selection for Best-of-N}} 

\author{
\centering{
Shubham Toshniwal,
Aleksander Ficek\protect\footnotemark[1],
Siddhartha Jain\protect\footnotemark[1],
Wei Du,
Vahid Noroozi,
Sadegh Mahdavi,
Somshubra Majumdar,
Igor Gitman
}
}

\begin{abstract}
\textbf{Abstract:}
Scaling test-time compute via parallel sampling can substantially improve LLM reasoning, but is often limited by Best-of-$N$ selection quality. Generative selection methods, such as GenSelect~\cite{toshniwal2024openmathinstruct2}, address this bottleneck, yet strong selection performance remains largely limited to large models. We show that small reasoning models can acquire strong GenSelect capabilities through targeted reinforcement learning.
To this end, we synthesize selection tasks from large-scale math and code instruction datasets by filtering to instances with both correct and incorrect candidate solutions, and train 1.7B-parameter models with DAPO~\cite{yu2025dapo} to reward correct selections.
Across math (AIME24, AIME25, HMMT25) and code (LiveCodeBench) reasoning benchmarks, our models consistently outperform prompting and majority-voting baselines, often approaching or exceeding much larger models.
Moreover, these gains generalize to selecting outputs from stronger models despite training only on outputs from weaker models.
Overall, our results establish reinforcement learning as a scalable way to unlock strong generative selection in small models, enabling efficient test-time scaling.

\end{abstract}

\begin{document}
\maketitle

\footnotetext[1]{Equal contribution.}

\begin{figure*}[ht]
    \hspace{-3mm}
    \centering
    \includegraphics[scale=0.6]{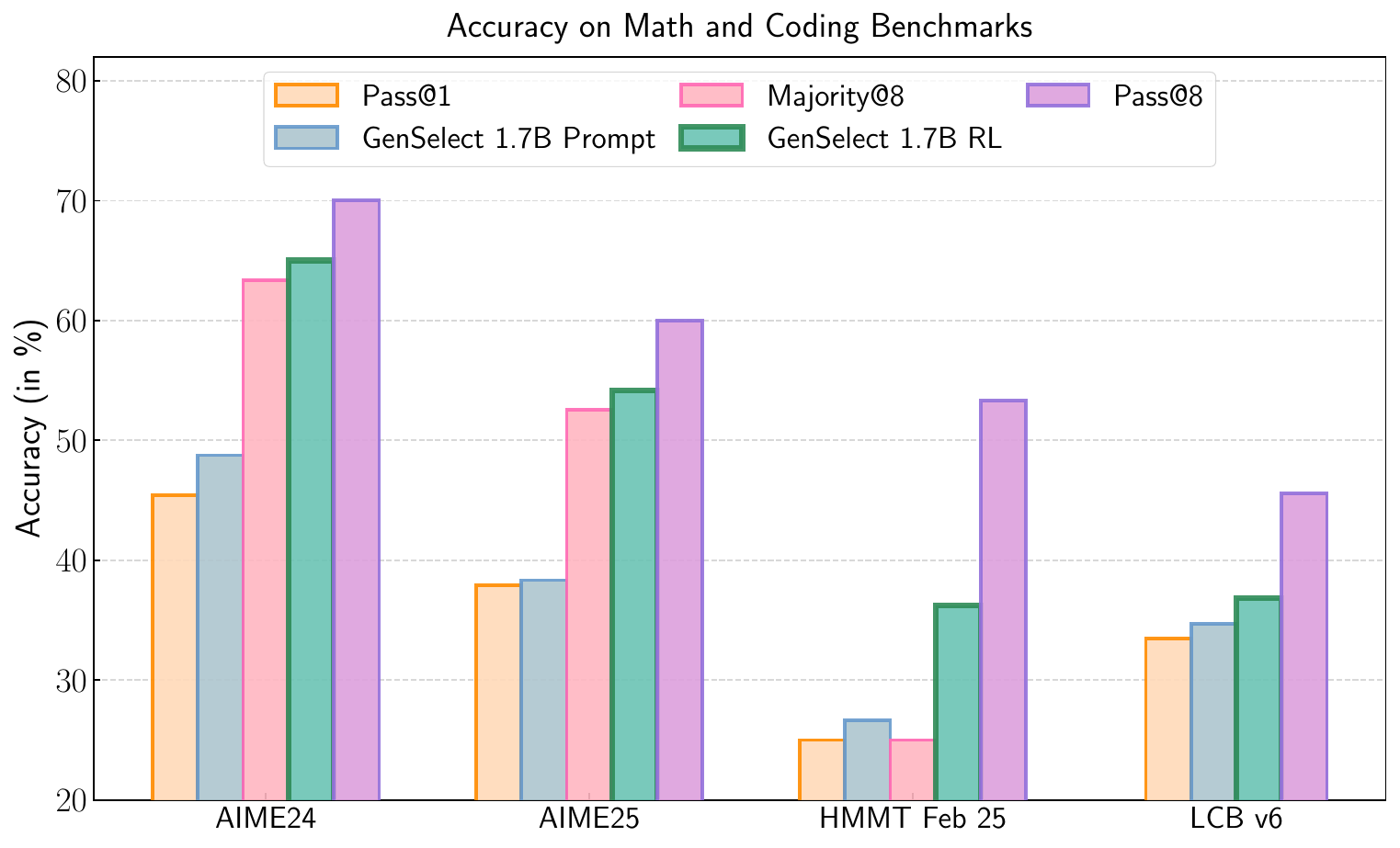}
    \caption{Accuracy of \qwenOneSevenB generations when selected using different strategies on math and coding benchmarks; RL-trained GenSelect consistently outperforms prompting and majority voting.}
    \label{fig:example_test}
\end{figure*}

\section{Introduction}

Test-time scaling has emerged as a powerful approach for improving the reasoning performance of large language models (LLMs)~\citep{openai2024openaio1card, deepseekai2025deepseekr1incentivizingreasoningcapability}. A simple and widely used strategy is \emph{parallel sampling}: generating multiple independent responses and selecting the best~\citep{brown2024largelanguagemonkeysscaling, snell2025scalingllmtesttimecompute}.
In practice, the effectiveness of parallel sampling is often limited by the quality of Best-of-$N$ selection (i.e., choosing the best answer among $N$ candidates), except for domains with automatic verifiers.

Existing Best-of-$N$ methods typically rely on shallow aggregation mechanisms, such as majority voting or self-consistency~\citep{wang2023selfconsistencyimproveschainthought}, or on external reward models that score each candidate independently~\citep{cobbe2021trainingverifierssolvemath, guo2025rewardreasoningmodel}. Recently, a new paradigm has emerged that uses reasoning models to perform aggregation over the $N$ responses, posing this aggregation as a reasoning problem~\citep{zhao2025majorityrightrltraining, qi2025learningreasonparallelsamples}. 
Among these, \texttt{GenSelect}~\citep{toshniwal2025genselectgenerativeapproachbestofn} prompts a reasoning model to explicitly select the best candidate (given $N$ candidate solutions, output the index of the selected candidate) and has demonstrated strong gains over prior methods.

In this work, we ask whether strong GenSelect behavior can be learned by \emph{small} reasoning models. We propose a targeted reinforcement learning framework for generative selection on math and coding tasks, where models are rewarded for selecting a correct solution among imperfect candidates. We construct candidate sets from problems on which the candidate generator achieves $\leq 50\%$ pass rate, ensuring nontrivial selection difficulty, and determine correctness using automatic verifiers. Using on-policy reinforcement learning, we train 1.7B-parameter reasoning models that consistently outperform prompting and majority-voting baselines and approach the selection performance of much larger 8B-scale models (\,$\sim$5$\times$ larger). Notably, these gains generalize to selecting higher-quality outputs produced by stronger generation models, suggesting that learned selection transfers beyond the training-time candidate distribution.

Our contributions are:
\begin{itemize}
    \item We introduce a reinforcement learning framework that trains small reasoning models to perform generative Best-of-$N$ selection.
    \item We construct math and coding selection tasks with automatically verified correctness and controllable selection difficulty.
    \item We show that 1.7B selectors substantially improve Best-of-$N$ performance and transfer to selecting candidates from stronger generators.
\end{itemize}

\begin{figure}[t]
  \centering

\begin{tcolorbox}[title={GenSelect Prompt}, colback=red!0, left=2pt,right=2pt,top=2pt,bottom=2pt]

{%
\ttfamily\normalsize
You will be given a problem followed by an enumerated list of \{num\_solutions\} candidate solutions. Your task is to systematically analyze these solutions to identify the best approach.

\vspace{0.3cm}

Problem: 
\{problem\}

\vspace{0.3cm}

Solutions: 
\{solutions\}

\vspace{0.3cm}
End your evaluation with exactly:
\vspace{0.3cm}

Judgment [IDX]
\vspace{0.3cm}

where IDX is the index 0-\{max\_idx\} of the best solution.
}

\end{tcolorbox}

  \caption{The GenSelect prompt used for selection in prompt baselines and RL training.}
  \label{fig:selection_prompt}
  \vspace{-0.2in}
\end{figure}

\section{Data Construction}

We construct GenSelect-style selection tasks for math and code using \qwenOneSevenB to generate candidate solutions and automatic verifiers (symbolic math checking or unit tests) to label correctness. We filter prompts to ensure they contain at least one correct solution while remaining nontrivial.

\subsection{Math}

We construct math selection data from \texttt{OpenMathReasoning}~\citep{moshkov2025aimo2winningsolutionbuilding} by selecting 37K problems on which the base model (\qwenOneSevenB) achieves a pass rate below 50\% under a symbolic verifier based on Math-Verify~\citep{Kydlicek_Math}. 

For each problem, we sample 2--16 candidate solutions from \qwenOneSevenB and assemble them into GenSelect-style prompts (\autoref{fig:selection_prompt}). We require at least one correct solution per prompt and cap the fraction of correct candidates at 50\% to keep the selection task nontrivial. We construct up to four prompts per problem by resampling candidate solution sets using the above recipe; for training efficiency, we discard prompts exceeding 16K tokens.

\subsection{Code}

For code, we use problems and test cases from \texttt{OpenCodeReasoning (OCR)}~\citep{ahmad2025opencodereasoningiisimpletesttime}. We generate candidate solutions with \qwenOneSevenB and execute them against the provided unit tests to obtain binary correctness labels.

We then construct GenSelect prompts analogously to the math setting (\autoref{fig:selection_prompt}), including up to 16 candidate solutions from \qwenOneSevenB and enforcing at least one correct solution with a 50\% correctness cap. We do this for both domains to bias the dataset toward harder problems and balance candidate sets to contain both correct and incorrect solutions, making the selection task nontrivial. We discard prompts exceeding 16K tokens and responses over 12K tokens.






\section{Experimental Setup}

\subsection{RL Training}
We train \qwenOneSevenB models for both math and code using the VeRL framework~\citep{sheng2025hybridflow} on NVIDIA H100 GPUs. We use the DAPO algorithm~\citep{yu2025dapo} with on-policy selection generation. 
We use a batch size of 128, 16 rollouts for math and 8 rollouts for code, a learning rate of $1\times10^{-6}$, and the AdamW optimizer. For sampling, we use temperature 1.5, top-$p$ 1.0, and a max output length of 16{,}384 tokens for both models.

\begin{table*}[!t]
\centering
  \small
  \setlength{\tabcolsep}{8pt}
  \renewcommand{\arraystretch}{1.2}
  \arrayrulecolor{black!65}

\begin{tabularx}{\textwidth}{l *{4}{>{\centering\arraybackslash}X}}
    \rowcolor{black!8}
    \toprule
      & AIME24 & AIME25 & HMMT25 & LCB v6 \\
    \midrule

    \rowcolor{black!8}
    \multicolumn{5}{l}{\textbf{Generator: \qwenOneSevenB}} \\
    \midrule
    Pass@1 & 45.42 ± 5.33 & 37.92 ± 6.16 & 25.00 ± 7.13 & 33.45 ± 1.08 \\
    Pass@8 & 70.00 & 60.00 & 53.33 & 45.59 \\
    \rowcolor{gray!8}
    Majority@8 & 63.33 & 52.56 & 25.00 & N/A \\
    \midrule
    \rowcolor{gray!8}
    GenSelect 1.7B Prompting & 48.75 ± 6.89 & 38.33 ± 5.04 & 26.67 ± 5.63 & 34.72 ± 0.71 \\
    GenSelect 4B Prompting & 61.25 ± 4.34 & 53.75 ± 4.15 & 34.17 ± 2.95 & 39.70 ± 0.81 \\
    GenSelect 8B Prompting & \underline{65.83} ± 2.36 & \underline{55.42} ± 3.05 & \underline{38.75} ± 3.05 & \underline{41.24} ± 0.67 \\
    \midrule
    \rowcolor{gray!8}
    GenSelect 1.7B RL Math & \textbf{65.00} ± 3.98 & \textbf{54.17} ± 2.36 & \textbf{36.25} ± 4.15 & 36.45 ± 0.86 \\
    \rowcolor{gray!8}
    GenSelect 1.7B RL Code & 57.08 ± 3.75 & 44.58 ± 3.54 & 26.25 ± 5.18 & \textbf{36.84} ± 0.44 \\

    \addlinespace[4pt]
    \midrule
    \rowcolor{black!8}
    \multicolumn{5}{l}{\textbf{Generator: Qwen3-4B}} \\
    \midrule
    Pass@1 & 72.08 ± 3.96 & 61.67 ± 3.98 & 41.67 ± 7.13 & 51.98 ± 0.58 \\
    Pass@8 & 86.67 & 83.33 & 60.00 & 66.30 \\
    \rowcolor{gray!8}
    Majority@8 & 80.00 & 72.64 & 50.00 & N/A \\
    \midrule
    \rowcolor{gray!8}
    GenSelect 1.7B Prompting & 74.58 ± 3.96 & 63.75 ± 3.75 & 42.92 ± 4.52 & 52.01 ± 1.09 \\
    GenSelect 4B Prompting & 80.00 ± 1.78 & 71.25 ± 2.48 & 50.00 ± 2.52 & 55.04 ± 0.88 \\
    GenSelect 8B Prompting & \underline{82.92} ± 2.14 & 72.50 ± 2.36 & 53.75 ± 2.78 & \underline{56.11} ± 1.09 \\
    \midrule
    \rowcolor{gray!8}
    GenSelect 1.7B RL Math & \textbf{80.83} ± 3.45 & \underline{\textbf{73.33}} ± 3.09 & \underline{\textbf{55.42}} ± 1.73 & \textbf{53.36} ± 0.81 \\
    \rowcolor{gray!8}
    GenSelect 1.7B RL Code & 77.50 ± 2.95 & 67.08 ± 3.30 & 47.92 ± 2.48 & 52.67 ± 0.34 \\

    \addlinespace[4pt]
    \midrule
    \rowcolor{black!8}
    \multicolumn{5}{l}{\textbf{Generator: Qwen3-8B}} \\
    \midrule
    Pass@1 & 76.25 ± 6.28 & 69.17 ± 4.27 & 40.42 ± 2.14 & 55.78 ± 0.69 \\
    Pass@8 & 90.00 & 86.67 & 56.67 & 66.96 \\
    \rowcolor{gray!8}
    Majority@8 & 83.81 & 76.67 & 50.00 & N/A \\
    \midrule
    \rowcolor{gray!8}
    GenSelect 1.7B Prompting & 78.75 ± 3.54 & 72.92 ± 3.75 & 46.25 ± 4.15 & 55.56 ± 0.67 \\
    GenSelect 4B Prompting & 83.33 ± 2.52 & 77.08 ± 2.14 & 52.08 ± 2.48 & 57.85 ± 0.87 \\
    GenSelect 8B Prompting & \underline{86.25} ± 2.14 & \underline{78.33} ± 2.52 & 55.83 ± 1.54 & \underline{58.26} ± 0.71 \\
    \midrule
    \rowcolor{gray!8}
    GenSelect 1.7B RL Math & \textbf{85.83} ± 2.36 & \underline{\textbf{78.33}} ± 1.78 & \underline{\textbf{56.25}} ± 1.18 & 56.33 ± 0.72 \\
    \rowcolor{gray!8}
    GenSelect 1.7B RL Code & 79.17 ± 2.36 & 74.58 ± 2.48 & 49.17 ± 2.95 & \textbf{56.51} ± 0.59 \\

    \bottomrule
  \end{tabularx}

  \captionsetup{skip=8pt}
  \caption{Accuracy in \% when selecting among $N$ candidate solutions generated by different models (Qwen3-1.7B/4B/8B). Pass@8 represents the maximum achievable selection accuracy while RL models compared with equivalent baselines are highlighted in gray. The best method among equivalently sized strategies is \textbf{bolded}, and the overall best method is \underline{underlined}.}
  \label{tab:selection_results}
\end{table*}
\FloatBarrier

\subsection{Evaluation}
For evaluation, we report math results on AIME24, AIME25, and HMMT25, and code results on the LiveCodeBench v6 split (August 2024--May 2025). All evaluations are performed with the NeMo-Skills framework~\citep{nemo-skills}. Prompting baselines use the same selection prompt employed during training (\autoref{fig:selection_prompt}). GenSelect results are averaged over 8 runs. Majority@8 corresponds to the accuracy obtained by selecting the most common answer (applicable only for math tasks), and Pass@8 defines the maximum achievable selection accuracy. 

\section{Results}

Table~\ref{tab:selection_results} reports the accuracy when applying our trained models to select among candidate sets generated by \qwenOneSevenB, \texttt{Qwen3-4B}, and \texttt{Qwen3-8B}.

\subsection{Results on \qwenOneSevenB Generations}

Table~\ref{tab:selection_results} (\qwenOneSevenB candidates) shows that reinforcement learning substantially improves GenSelect performance over pass@1, majority voting, and GenSelect prompting with \qwenOneSevenB. In math, the RL-trained 1.7B model surpasses the out-of-the-box performance of Qwen3-4B and approaches that of Qwen3-8B across AIME24, AIME25, and HMMT25. In code, the RL-trained 1.7B model consistently outperforms same-size prompting baselines but does not exceed Qwen3-4B or Qwen3-4B GenSelect prompting.

We observe that GenSelect prompting with \qwenOneSevenB underperforms its 4B and 8B counterparts noticeably despite identical prompts. While lower values are expected, we hypothesize this is also due to the smaller 1.7B parameter count, which may cause the model to overfit to typical solution-generation tasks and fail to generalize to the GenSelect instructions. With a small amount of RL training, we enable the model to learn solution selection at a level comparable to an 8B-parameter model.

\subsection{Results on Higher Quality Generations}

\begin{figure*}
    \centering
    \includegraphics[width=1.00\linewidth]{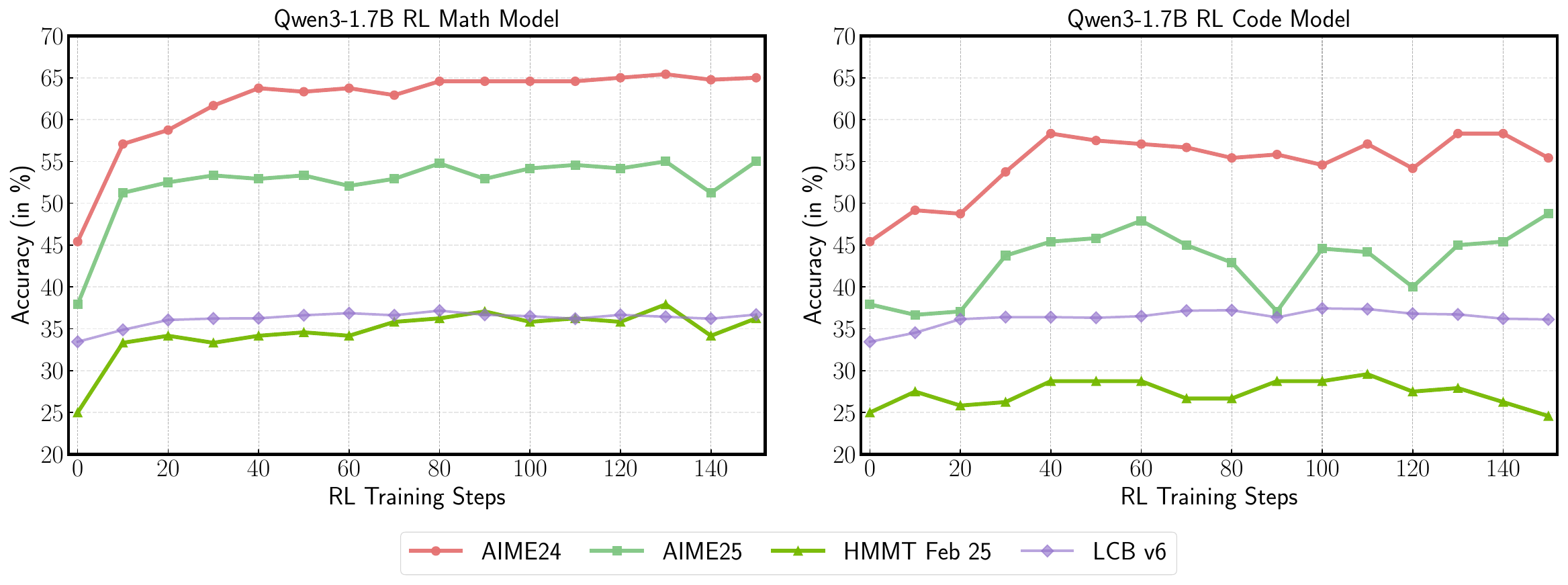}
    \caption{RL training dynamics of \qwenOneSevenB for GenSelect on math (left) and code (right) selection tasks. Selection accuracy improves rapidly and stabilizes for math, while training on code shows noisier dynamics and earlier saturation, reflecting domain-dependent differences in selection difficulty.}
    \label{fig:twofigs}
\end{figure*}

Table~\ref{tab:selection_results} (Qwen3-4B and Qwen3-8B candidates) demonstrates that the gains from reinforcement learning generalize to higher-quality candidate sets. When selecting from Qwen3-4B and Qwen3-8B generations, our RL-trained math models match or exceed the out-of-the-box selection performance of Qwen3-8B on AIME25 and HMMT25, while remaining competitive on AIME24 and LiveCodeBench.

These results are particularly notable because selection over Qwen3-4B and Qwen3-8B generations is off-policy relative to the \qwenOneSevenB-generated candidates used during training. This indicates that reinforcement learning enables the model to learn a transferable notion of solution quality rather than overfitting to a specific generation model. Across all generation qualities, GenSelect models trained with RL outperform majority voting and equivalently sized prompting baselines. This establishes reinforcement learning as an effective strategy for improving parallel thinking and one that generalizes to different generation models.


\subsection{Math and Coding Model Differences}

Performance gains from reinforcement learning are consistently larger for math than for code. While RL-trained code models outperform GenSelect prompting with Qwen3-1.7B, they remain behind 4B and 8B prompting baselines on LiveCodeBench. We also observe asymmetric cross-domain transfer: training on math selection data improves code selection, whereas training on code yields limited improvements on math, consistent with prior findings~\citep{chen2025acereasonnemotronadvancingmathcode}. \autoref{fig:twofigs} shows the accuracy after training of increments of 10 steps steadily improving the math model on both math and coding domains. Alternatively, the coding model demonstrates a similar pattern for LCB but marginal and less stable improvements on math benchmarks.

We attribute the smaller gains in code to two factors. First, correctness signals derived from unit tests are noisier and less exhaustive than symbolic math verification, weakening the reinforcement signal: a buggy program can pass a limited test suite (false positives), and otherwise-correct solutions can fail due to corner cases not covered by the tests or to runtime/formatting issues. Second, code selection requires broader semantic coverage than math---including algorithmic structure, edge-case handling, complexity constraints, and implementation details---making comparative reasoning more difficult than in math, where correctness is often more sharply defined~\citep{wang2025codecontestshighqualitytestcase}.

\section{Related Work}

Test-time compute has emerged as a highly effective method for improving LLM reasoning, and has become a central theme across recent work on scaling reasoning capabilities. Early work on self-consistency~\citep{wang2023selfconsistencyimproveschainthought} demonstrated that sampling multiple reasoning paths and aggregating them via majority voting can substantially improve accuracy. Subsequent surveys and scaling studies~\citep{zhang2025surveytesttimescalinglarge, zhao2025samplescrutinizescaleeffective} have systematized this direction, identifying test-time compute as a promising and complementary alternative to scaling model size.

\subsection{Reward Models}

Reward models have proven effective for rating and selecting candidate solutions, with multiple works highlighting their benefits in mathematical and coding tasks~\citep{liu2025acemathadvancingfrontiermath, zeng2025acecoderacingcoderrl}. To further exploit the advantages of test-time compute, recent efforts have investigated generative approaches: verifiers framed as next-token predictors~\citep{zhang2025generativeverifiersrewardmodeling}, reward models trained on reasoning preferences~\citep{mahan2024generativerewardmodels, guo2025rewardreasoningmodel}, and models that reason via Chain-of-Thought before scoring. Other directions include pairwise tournament-style judging \citep{liu2025pairjudgermperformbestofn}, generating test cases as verifiers \citep{ficek2025scoringverifiersevaluatingsynthetic}, and reinforcement-learning verifiers for competitive math proofs \cite{shi2025heimdalltesttimescalinggenerative}. Further refinements leverage critique fine-tuning to enhance scoring \cite{wang2025critiquefinetuninglearningcritique, ahmad2025opencodereasoningiisimpletesttime} and internal model confidence signals to filter reasoning traces \cite{fu2025deepthinkconfidence}.

\subsection{Parallel Reasoning Selection} 
Parallel reasoning methods extend beyond scoring to structured comparison, aggregation, and refinement over multiple candidate solutions. GenSelect~\citep{toshniwal2025genselectgenerativeapproachbestofn} reframes Best-of-$N$ as a generative selection problem, using a reasoning model to explicitly compare and select among candidates. Related approaches explore self-refinement and iterative improvement during training~\citep{wang2025learningrefineselfrefinementparallel}, as well as aggregation over parallel outputs using learned reasoning policies~\citep{qi2025learningreasonparallelsamples, zhao2025majorityrightrltraining}.

In this work, we focus exclusively on \emph{selection} rather than synthesis. This choice is motivated by the strong empirical performance of GenSelect~\citep{toshniwal2025genselectgenerativeapproachbestofn} and findings by \citet{qi2025learningreasonparallelsamples} showing that synthesized outputs often reduce to copying one of the inputs. 

While synthesis strategies are effective, in this work we show that selection can be equally successful. In addition, selection can be substantially easier to train with reinforcement learning: since rewards are defined over selected candidates, correctness labels can be precomputed for the entire candidate set. In contrast, synthesis-based aggregation produces new outputs that typically require re-running verification (or reward-model evaluation), which can be costly or difficult to integrate in complex RL environments. We extend the work by \citet{toshniwal2025genselectgenerativeapproachbestofn} by proposing training strategies to enhance the GenSelect performance of small reasoning models.


\section{Conclusion}

We study reinforcement learning for GenSelect in small reasoning models on the Best-of-$N$ selection problem in math and code. Training a 1.7B-parameter model directly for generative selection consistently outperforms majority voting and GenSelect prompting, in several cases exceeding the performance of much larger models. The learned selection behavior generalizes beyond the training-time candidate distribution, enabling effective selection among candidates generated by the frozen 1.7B base model at evaluation time. Reinforcement learning yields larger and more reliable gains in math than in code, with stronger cross-domain transfer when training on math. Overall, our results show that training for selection is a useful complementary approach to test-time scaling, enabling small models to exhibit powerful aggregation behavior.

\bibliographystyle{plainnat}  
\bibliography{paper}  

\appendix

\end{document}